\documentclass[sn-mathphys-num]{sn-jnl}

\usepackage{graphicx}%
\usepackage{multirow}%
\usepackage{amsmath,amssymb,amsfonts}%
\usepackage{amsthm}%
\usepackage{mathrsfs}%
\usepackage[title]{appendix}%
\usepackage{xcolor}%
\usepackage{textcomp}%
\usepackage{manyfoot}%
\usepackage{booktabs}%
\usepackage{algorithm}%
\usepackage{algorithmicx}%
\usepackage{algpseudocode}%
\usepackage{listings}%

\usepackage{anyfontsize}
\usepackage{ragged2e} 
\usepackage{multirow}
\usepackage{array}

\usepackage{rotating}

\DeclareMathOperator*{\argmin}{argmin}

\usepackage{titlesec}
\titleformat{\paragraph}[runin]
  {\normalfont\normalsize\bfseries} % font style
  {}{0pt}{} % no numbering, no extra spacing
\titlespacing*{\paragraph}
  {0pt}{3.25ex plus 1ex minus .2ex}{1em} % left indent, before-skip, after-skip

\begin{document}

\title[Penalty Learning for Changepoint Detection via Automatic Feature Extraction using RNN]{Penalty Learning for Changepoint Detection via Automatic Feature Extraction using RNN}

\author*[1]{\fnm{Tung} \sur{L Nguyen}}\email{tln229@nau.edu}
\author[2]{\fnm{Toby} \sur{Dylan Hocking}}\email{toby.dylan.hocking@usherbrooke.ca}

\affil*[1]{\orgdiv{School of Informatics, Computing, and Cyber Systems}, \orgname{Northern Arizona University}, \orgaddress{\street{S San Francisco}, \city{Flagstaff}, \postcode{86011}, \state{Arizona}, \country{USA}}}
\affil[2]{\orgdiv{Département d'informatique}, \orgname{Université de Sherbrooke}, \orgaddress{\street{Sherbrooke QC J1K 2R1}, \city{Quebec}, \country{Canada}}}

\abstract{Changepoint detection identifies significant shifts in data sequences, making it important in areas like finance, genetics, and healthcare.
The Optimal Partitioning algorithms efficiently detect these changes, using a penalty parameter to limit the changepoints number.
Determining the optimal value for this penalty can be challenging. 
Traditionally, this process involved manually extracting statistical features, such as sequence length or variance to make the prediction.
This study proposes a novel approach that uses recurrent neural networks (RNN) to learn this penalty directly from raw sequences by automatically extracting features.
Experiments conducted on 20 benchmark datasets show that this novel method generally outperforms traditional ones in changepoint detection accuracy.}


\maketitle

\section{Introduction} \label{sec:intro}
\begin{figure}[t]
\includegraphics[width=0.99\textwidth]{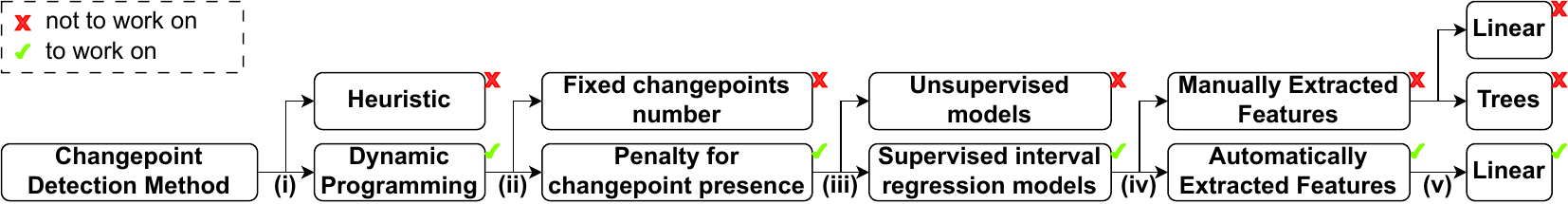}
\caption{Literature review of this study: (i) Selecting a detection method, (ii) Choosing a regularization method to control the number of changepoints, (iii) Selecting a model type to predict the penalty, (iv) Set up the penalty prediction problem, and (v) Choosing the model architecture.}
\label{fig:literature}
\end{figure}

Changepoint detection (other words, partitioning or segmentation) is important for identifying abrupt changes in data in various domains, including finance \citep{app:finance}, healthcare \citep{cancer}, network security \citep{app:network}, and environmental science \citep{app:climate}. This importance has driven the development of numerous methods over the past decades.
The literature review, gap in existing methods, and study contributions are based on the diagram in Figure \ref{fig:literature}.

\paragraph{Changepoint Detection Methods - Fig. \ref{fig:literature} (i).}
The methods can be divided into Heuristic and Dynamic Programming (DP).
% Pure DL and DP approaches take different perspectives.
% DP methods treat the problem as an optimization task, aiming to find the optimal partition, whereas DL methods frame it as a binary classification problem, predicting whether each candidate point is a changepoint. 
% So in terms of the underlying optimization, DP methods are optimal, while DL methods are heuristic.
% This category division approach is similar to that of Paul Fearnhead, the author of \citep{review:pelt} (a DP method), and \citep{review:li24automatic} (a DL method) where he classifies changepoint detection methods into traditional and neural network–based.
Heuristic methods include deep learning (DL)-based approaches inspired by the statistical Cumulative Sum (CUSUM) method \citep{review:cusum, review:james1987tests}, including recent works \citep{review:li24automatic, review:ermshaus2023clasp} employ multilayer perceptrons (MLP) or k-nearest neighbors (KNN) as classification sliding windows to identify changepoints. 
However, these methods rely on local neighbors rather than the entire sequence and are overly complex for univariate data.
In contrast, DP is well-suited for this context.

\paragraph{Regularization - Fig.~\ref{fig:literature} (ii).} The DP algorithms rely on a single hyperparameter to limit the changepoints number, which can be either: (a) a fixed changepoints number or (b) a penalty to penalize the changepoint presence.
Fixed changepoints number allow DP algorithms \citep{review:binseg, review:bai2003computation, review:pelt, rigaill2015pruned} to locate them, but the changepoints number is rarely predetermined.
More commonly used, penalty-based DP algorithms, such as Optimal Partitioning (OPART) \citep{review:opart} and its variants—Pruned Exact Linear Time (PELT) \citep{review:pelt}, Functional Pruning Optimal Partitioning (FPOP) \citep{review:fpop}, and Labeled Optimal Partitioning (LOPART) \citep{review:lopart}—are widely regarded as the most effective.
PELT and FPOP prune candidate changepoints efficiently, producing identical partitions as OPART, while LOPART extends OPART by incorporating predefined labels (the expected number of changepoints within specific location ranges) or defaulting to OPART when labels are absent.
OPART is formulated as an optimization problem shown below:
\paragraph{OPART Problem.} Given a sequence $\mathbf{d} \in \mathbb{R}^N$ and penalty $\lambda \geq 0$,
find the model parameter vector $\mathbf{m} \in \mathbb{R}^N$ that minimizes the following cost function:
\begin{equation*} \label{eq:opart_loss}
C(\textbf{d}, \textbf{m}) + \lambda \sum_{i=1}^{N-1} \mathbf{1}(m_i \neq m_{i+1}),
\end{equation*}
\noindent where $\mathbf{1}(z)$ equals 1 when $z$ is True and 0 otherwise.
The $C(\textbf{d}, \textbf{m})$ formula depends on the type of change.
For example, if the changes is the mean of normal distribution, 
then $C(\mathbf{d}, \mathbf{m}) = \sum_{i=1}^{N} (d_i - m_i)^2,$ 
or for changes in the variance, 
$C(\mathbf{d}, \mathbf{m}) = \sum_{i=1}^{N} \big((d_i - m_i)^2 - \sigma_i^2\big)^2,$ where \(\sigma_i^2\) is the variance of the segment corresponding to \(m_i\).

\begin{figure}[t]
\vskip 0.2in
\begin{center}
\centerline{\includegraphics[width=0.8\textwidth]{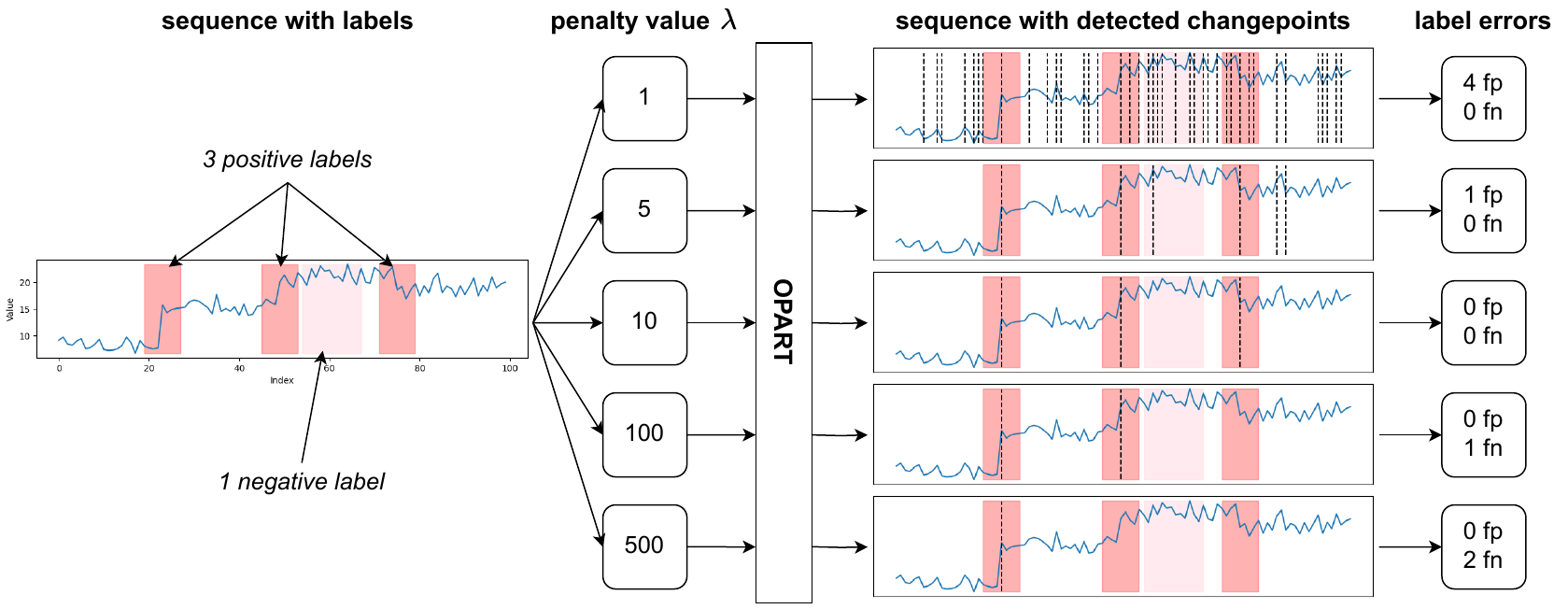}}
\caption{Example of how $\lambda$ affect the OPART changepoints. The sequence contains 4 labels: 3 positive (1 changepoint regions) and 1 negative (no changepoint region). Two types of errors are considered: false positives (fp), where extra changepoints are detected in any labels, and false negatives (fn), where no changepoint is detected in a positive label. In this examlpe, $\lambda=10$ is optimal.}
\label{fig:lambda}
\end{center}
\vskip -0.2in
\end{figure}

In OPART algorithms, the penalty plays a key role.
A higher penalty imposes a stronger penalty on changepoint presence, leading to fewer changepoints, see Figure \ref{fig:lambda}.
This study aims to predict that penalty to improve the changepoint detection accuracy. 
From here on, that conceptual penalty will be referred to as $\lambda$.

\paragraph{Existing $\lambda$ prediction models - Fig.~\ref{fig:literature} (iii).}
Various models exist for predicting $\lambda$, including unsupervised \citep{bic, aic, lavielle05} and supervised such as linear \citep{rigaill:13} and tree(s) \citep{mmit, barnwal2022survival, nguyen2025intervalregressioncomparativestudy}.
But all of these models rely on a set of manually extracted sequence features, yet the number of possible features is infinite, this process can result in the inclusion of many irrelevant features, and the potential omission of useful ones.

\paragraph{Contribution - Fig.~\ref{fig:literature} (iv).} This study proposes a novel method utilizing a recurrent network to automatically extract relevant features from sequences, which are then used to predict $\lambda$.
Experimental results on benchmark datasets show that this method outperforms existing methods in terms of OPART accuracy in most cases.

\paragraph{Clarification.} The proposed model is \textit{not} a heuristic neural network–based changepoint detection method.
Instead, it focuses on learning the penalty for OPART, a classical parametric changepoint algorithm grounded in maximum likelihood inference.

\paragraph{Reproducibility.} To ensure transparency, the code used for this study is accessible at: \url{https://github.com/lamtung16/ML_Changepoint_Detection_epigenomic_rnn}.

\section{Problem Setting and Previous Work} \label{sec:previous}
\subsection{Problem Setting}
\paragraph{Study supervision.} After visually inspecting the signal and noise patterns in the data, an expert labels the region by specifying its start and end positions, along with the expected number of changes (usually 0 or 1).
Each sequence is assigned multiple labels, see the example in Figure \ref{fig:lambda} of a sequence having four labels.

\paragraph{The $\lambda$ prediction models use interval regression.} This study aims to predict the optimal $\lambda$ for each sequence and then apply OPART with the predicted $\lambda$ to detect its changepoints.
For each labeled sequence, the optimal $\lambda$ is \textit{not a single value}, but rather is an \textit{interval} $[\lambda_l, \lambda_u]$.
For example, in Figure \ref{fig:lambda}, $\lambda = 10$ is the optimal value, as it generates changepoints that align with expert labels; however, there exist $a, b \geq 0$ such that values of $\lambda$ within $[10-a, 10+b]$ could produce the same result, so $[10-a, 10+b]$ is the \textit{target interval} of that sequence.
The algorithm for generating the target interval for each labeled sequence is detailed by \citet{rigaill:13}. 
This problem, where the model predicts the value that falls within the target interval, is known as \textit{interval regression}.
In interval regression, there are 4 types of target intervals $[y_l, y_u]$: uncensored $[ -\infty < y_l = y_u < \infty ]$, interval-censored $[ -\infty < y_l < y_u < \infty ]$, left-censored $[ -\infty = y_l < y_u < \infty ]$ and right-censored $[ -\infty < y_l < y_u = \infty ]$.
In this study, each labeled sequence is associated with an interval $[\lambda_l, \lambda_u]$ where $0 \leq \lambda_l \leq \lambda_u \leq \infty$. 

\subsection{Previous Work}
This subsection focuses on the process of training supervised $\lambda$ prediction models.
% From this point onward, the prediction of $\lambda$ is denoted by $\hat{\lambda}$ to distinguish it from the conceptual OPART penalty $\lambda$.
Before going into the details, some noteworthy models are highlighted, which, although not applicable to this study, are still relevant. 
Additionally, some unsupervised models, which have served as inspiration for supervised models, are mentioned.

\paragraph{Noteworthy.} Adaptive Linear Penalty INference (ALPIN) \citep{truong17} is a method for this supervision special case, where every point is a label, equivalent to label regions having length 1.
However, since this study uses a more relaxed supervision with varying label region lengths, ALPIN is not applicable.
The Accelerated Failure Time (AFT) model \citep{aft} and its variants \citep{cai2009regularized, huang2006regularized, quinlan1986induction, svm}, which are designed for censored outcomes, can be considered.
Supervised changepoint detection setups involve interval-censored targets, whereas traditional AFT models are restricted to uncensored or right-censored intervals, making them not directly applicable. 
% An important distinction is that in AFT models, the left-censored interval is defined as $ (0 = \lambda_l < \lambda_u < \infty) $, which is different from this study's definition.

\paragraph{Unsupervised.} These models predicts $\lambda$ based on some sequence features.
Let $\hat{\lambda}$ denote a prediction of $\lambda$.
Let $N$ denote the sequence length, and $\sigma$ its standard deviation, with examples such as Baysian Information Criterion (BIC) \citep{bic} predicts $\hat{\lambda} = \sigma^2 \log N$.
Akaike's Information Criterion (AIC) \citep{aic} predicts $\hat{\lambda} = 2p$, where $ p $ represents a sequence feature (e.g., standard deviation, variance).
\citet{lavielle05} does not provide a closed-form formula but considers $\hat{\lambda}$ to be dependent on both $ N $ and $ \sigma $.

\paragraph{Supervised.} The rest of this section will detail the training process of three supervised machine learning models: Maximum Margin Interval Regression \citep{rigaill:13}, a linear model; Maximal Margin Interval Tree (MMIT) \citep{mmit}; and AFT in XGBoost \citep{barnwal2022survival}, which utilizes a tree-based framework (although AFT in XGBoost is an AFT model, it is applicable due to its ability to handle all types of non-negative intervals).
These models divided the training process into 2 sequential independent steps:
\begin{itemize}
    \item Step 1: Extract features vector $\mathbf{x}$ from the sequence.
    \item Step 2: Train a model $g(\cdot)$ takes into $\mathbf{x}$ to predict $\hat{\lambda} = g(\mathbf{x})$
\end{itemize}

\subsubsection{Step 1: Feature Extraction}
The manual feature extraction process, detailed in \cite{rigaill:13}, involves the following steps:
\begin{itemize}
    \item Starting from the sequence $\mathbf{d} = [d_1, d_2, \dots, d_N]$ with the mean value $\bar{d} = \frac{1}{N} \sum_{i=1}^N d_i$, 2 additional sequences (residuals and differences) are generated:
    \begin{align*}
        \mathbf{d}_{\text{res}} &= [d_1 - \bar{d}, \dots, d_N - \bar{d}] \in \mathbb{R}^N \\
        \mathbf{d}_{\text{diff}} &= [d_2 - d_1, \dots, d_N - d_{N-1}] \in \mathbb{R}^{N-1}
    \end{align*}

    \item Apply 3 transformations (identity, absolute, square) to each of 3 vectors, resulting in 9 transformed vectors.
    \item Compute 8 statistics (sum, mean, standard deviation, and quantiles at 0\%, 25\%, 50\%, 75\%, 100\%) from each transformed vector, yielding a vector of length 72.
    \item Include the sequence length as an additional feature, forming a vector of length 73.
    \item Apply 5 transformations (identity, square root, log, log-log, square) to each feature, resulting in up to 365 features.
\end{itemize}
The values caused by invalid operations (e.g., log of a negative number), is ignored.

\subsubsection{Step 2: Model Training}
After obtaining the feature vector $\mathbf{x}$ (up to 365 features), each instance $\{\mathbf{d}, [\lambda_l, \lambda_u]\}$ becomes $\{\mathbf{x}, [\lambda_l, \lambda_u]\}$, a selected model $g$ is trained to predict $\hat{\lambda} = g(\mathbf{x})$.

\paragraph{AFT in XGBoost.} The model is formulated as:
\begin{equation*}
    \hat{\lambda} = \mathbf{T}(\mathbf{x}) + \varepsilon
\end{equation*}
Here, $\mathbf{T}(\mathbf{x})$ denotes the ensemble of trees for feature vector $\mathbf{x}$, and $\varepsilon$ is a random error. 
The model is trained by minimizing the negative log-likelihood over all train instances:
\begin{equation} \label{eq:aft_loss}
    l_{AFT}\Big(\hat{\lambda}, [\lambda_l, \lambda_u]\Big) = -\log \Big[ F_Z\big( \frac{\log\lambda_u - \log\hat{\lambda}}{\sigma} \big) - F_Z\big(\frac{\log\lambda_l - \log\hat{\lambda}}{\sigma}\big) \Big]
\end{equation}

\noindent where $\sigma > 0$ is the scale parameter.
The cumulative distribution function $F_Z(z)$ depends on the chosen distribution of $\varepsilon$:
\begin{itemize}
    \item normal: \phantom{00}   $F_Z(z) = \frac{1}{2}\big( 1 + \text{erf}(\frac{z}{\sqrt{2}}) \big)$
    \item logistic: \phantom{00} $F_Z(z) = \frac{e^z}{1+e^z}$
    \item extreme: \phantom{0} $F_Z(z) = 1 - e^{-\exp z}$
\end{itemize}
The model is trained using the gradient boosting algorithm \citep{gradient_boosting} to minimize the loss value \eqref{eq:aft_loss} over the training dataset with $S$ instances $\{\mathbf{x}^i, [\lambda_l^i, \lambda_u^i]\}_{i=1}^S$ given the distribution of $\varepsilon$ and the scale parameter $\sigma$:
\begin{equation*}
    \sum_{i = 1}^S l_{AFT}\Big(\hat{\lambda}^i, [\lambda_l^i, \lambda_u^i]\Big) = -\sum_{i=1}^S \log \Big[ F_Z\big( \frac{\log\lambda_u^i - \log\mathbf{T}(\mathbf{x}^i)}{\sigma} \big) - F_Z\big(\frac{\log\lambda_l^i - \log\mathbf{T}(\mathbf{x}^i)}{\sigma}\big) \Big]
\end{equation*}

% \begin{figure}[!t]
% \vskip 0.2in
% \begin{center}
% \centerline{\includegraphics[width=0.6\textwidth]{figure.2_loss_function.pdf}}
% \caption{Example about the loss value of two loss functions is determined by comparing the prediction to the target.
% In the top plot, the Squared Error yields a value of 0 when the prediction $\hat{y}$ reaches the target point $y$. The bottom plot illustrates the Squared Hinge Error, where the loss value reaches 0 when the prediction $\hat{y}$ falls within the interval $[y_l + \epsilon, y_u - \epsilon]$}
% \label{fig:loss_func}
% \end{center}
% \vskip -0.2in
% \end{figure}

\paragraph{Linear.} This model, known as Maximum Margin Interval Regression, transforms the problem to predicting $\log \hat{\lambda}$ instead of $\hat{\lambda}$, as $\log \hat{\lambda}$ can be any real value, unlike the non-negative $\hat{\lambda}$. The model expresses the output as a linear combination of inputs:
\begin{equation} \label{eq:linear_prediction}
    \log \hat{\lambda} = \mathbf{x} \cdot \beta + \beta_0
\end{equation}
where $\beta$ is the parameter vector, $\mathbf{x} \cdot \beta$ is the dot product of $\mathbf{x}$ and $\beta$, and $\beta_0$ is the bias term.
Its simplified version, used in \cite{review:lopart}, predicts $\log \hat{\lambda} = \beta_1 \times \log \log N + \beta_0$, where $N$ is the sequence length.

This model is trained by minimizing a loss function, the Hinge Error, defined by \citet{rigaill:13}, it achieves 0 loss when the prediction falls within the target interval.
The error value between prediction $\log \hat{\lambda}$ and target interval $[\log \lambda_l, \log \lambda_u]$ is: 

$$l\Big(\log \hat{\lambda}, [\log \lambda_l, \log \lambda_u]\Big) = 
\begin{cases} 
    (\log \lambda_l - \log \hat{\lambda} + \epsilon)^2 & \mbox{if } \log \hat{\lambda} < \log \lambda_l + \epsilon, \\
    0 & \mbox{if } \log \lambda_l + \epsilon \leq \log \hat{\lambda} \leq \log \lambda_u - \epsilon, \\
    (\log \hat{\lambda} - \log \lambda_u + \epsilon)^2 & \mbox{if } \log \hat{\lambda} > \log \lambda_u - \epsilon.
\end{cases}$$

\noindent where $\epsilon \geq 0$ is the margin length ($\epsilon=0$ by default).
The Hinge Error becomes Square Error when $\lambda_l = \lambda_u$ and $\epsilon = 0$ (or $\epsilon = \frac{\log\lambda_l+\log\lambda_u}{2}$), or in other word, target interval becomes target point.
\citet{mmit} proposed a similar loss function:
$$l\Big(\log \hat{\lambda}, [\log \lambda_l, \log \lambda_u]\Big) = 
\begin{cases} 
    |\log \lambda_l - \log \hat{\lambda} + \epsilon| & \mbox{if } \log \hat{\lambda} < \log \lambda_l + \epsilon, \\
    0 & \mbox{if } \log \lambda_l + \epsilon \leq \log \hat{\lambda} \leq \log \lambda_u - \epsilon, \\
    |\log \hat{\lambda} - \log \lambda_u + \epsilon| & \mbox{if } \log \hat{\lambda} > \log \lambda_u - \epsilon.
\end{cases}$$
Using the ReLU function, the general loss function can be written in this short form:
\begin{equation} \label{eq:loss_function}
    l(\log \hat{\lambda}, [\log \lambda_l, \log \lambda_u]) = \Big( \text{ReLU}(\log \lambda_l - \log \hat{\lambda} + \epsilon)\Big)^p + \Big(\text{ReLU}(\log \hat{\lambda} - \log \lambda_u + \epsilon) \Big)^p,
\end{equation}where $\epsilon \geq 0$ is the margin length and $p \in \{1, 2\}$ is the loss type.

To estimate the parameters $\beta$ and $\beta_0$ in \eqref{eq:linear_prediction}, the Fast Iterative Shrinkage-Thresholding Algorithm (FISTA) \citep{fista} with respect to $\beta$ and $\beta_0$ is employed to minimize the Hinge Error \eqref{eq:loss_function} over the train dataset having $S$ instances $\{\mathbf{x}^i, [\log \lambda_l^i, \log \lambda_u^i]\}_{i=1}^S$:
\begin{equation*}
    \sum_{i = 1}^S l\Big(\log \hat{\lambda}^i, [\log \lambda_l^i, \log \lambda_u^i]\Big) = \sum_{i = 1}^S l\Big(\mathbf{x^i} \cdot \beta + \beta_0, [\log \lambda_l^i, \log \lambda_u^i]\Big)
\end{equation*}

\begin{figure}[t]
\vskip 0.2in
\begin{center}
\centerline{\includegraphics[width=0.9\textwidth]{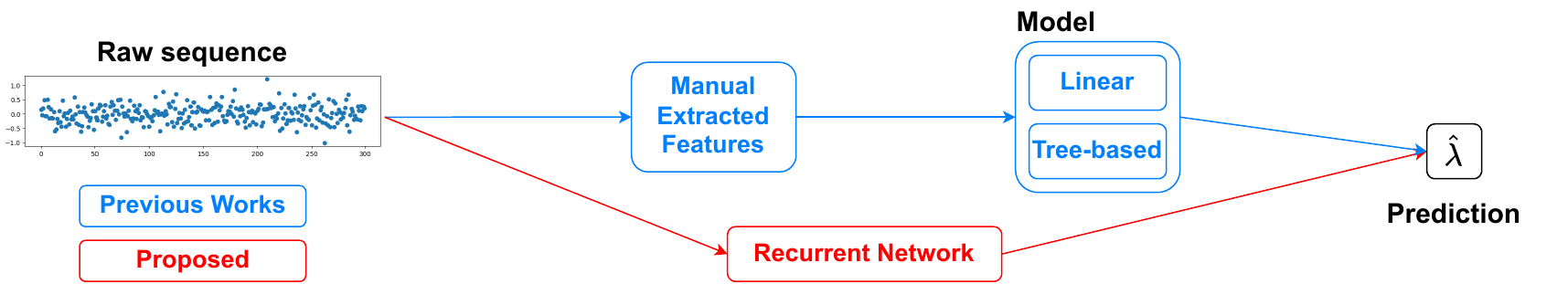}}
\caption{Diagram of the methods - Instead of performing feature extraction and the learning model separately, the proposed method uses a recurrent network to combine both into a single operation.}
\label{fig:proposed}
\end{center}
\vskip -0.2in
\end{figure}

\paragraph{Tree.} MMIT introduced by \citet{mmit} is applicable for $\lambda$ prediction as an interval regression model. 
The MMIT model is:
\[
\log \hat{\lambda} = \mathcal{T}(\mathbf{x})
\]
The tree $\mathcal{T}$ architecture follows the same structure as CART (Classification And Regression Tree) \citep{cart}. 
The only difference lies in the regression value for each leaf which is a set of $Q$ instances $\{\mathbf{x}^i, [\log \lambda_l^i,\log \lambda_u^i]\}_{i=1}^Q$. Instead of taking the mean of all targets like CART does, MMIT chooses a constant value $c$ that minimizes the mean Hinge Error \eqref{eq:loss_function} between this constant and the target intervals:
\begin{equation*}
    \argmin_c \frac{1}{Q} \sum_{i=1}^{Q} l\big(c, [\log \lambda_l^i, \log \lambda_u^i]\big), \quad c \in \mathbf{R}
\end{equation*}
\noindent A heuristic algorithm is used to speed up the estimation of the mean value for each leaf by considering a finite set of candidates, rather than solving for the optimal $c$:
\begin{align*}
    &\argmin_c \frac{1}{Q} \sum_{i=1}^{Q} l\big(c, [\log \lambda_l^i, \log \lambda_u^i]\big), 
    \\
    &\text{where }c \in \{\log \lambda_l^i, \log \lambda_u^i, \frac{\log \lambda_l^i + \log \lambda_u^i}{2}|\log \lambda_l^i > -\infty, \log \lambda_u^i < \infty\}_{i = 1}^Q
\end{align*}

\noindent The training procedure for determining the optimal tree architecture follows the same operations as in CART (binary optimization) to minimize the Hinge Error \eqref{eq:loss_function} over the train dataset having $S$ instances $\{\mathbf{x}^i, [\log \lambda_l^i, \log \lambda_u^i]\}_{i=1}^S$:
\begin{equation*}
    \sum_{i = 1}^S l\Big(\log \hat{\lambda}^i, [\log \lambda_l^i, \log \lambda_u^i]\Big) = \sum_{i = 1}^S l\Big(\mathcal{T}(\mathbf{x}^i), [\log \lambda_l^i, \log \lambda_u^i]\Big)
\end{equation*}

\begin{table}[t]
\begin{tabular}{lccccc}
\toprule
\multicolumn{6}{c}{Recurrent core: $h_t = f(d_{[t-k+1:t]}; h_{t-1}) \in \mathbb{R}^m$} \\ \toprule
\textbf{sequence feature} & $k$ & $m$ & $h_{k-1}$ & $h_t$ & \textbf{used in}
\\
\midrule
length & $k \in \mathbf{N^+}$ & 1 & $k-1$ & $h_{t-1} + 1$ & \citep{bic, aic, lavielle05, mmit, barnwal2022survival, rigaill:13, review:lopart, opart_pen_mlp} \\
\cmidrule(lr){1-6}
mean & 1 & 1 & 0 & $h_{t-1} + \frac{d_t - h_{t-1}}{t}$ & \citep{mmit, barnwal2022survival, rigaill:13} \\
\cmidrule(lr){1-6}
min & 1 & 1 & $\infty$ & $\min(d_t, h_{t-1})$ & \citep{mmit, barnwal2022survival, rigaill:13} \\
\cmidrule(lr){1-6}
max & 1 & 1 & $-\infty$ & $\max(d_t, h_{t-1})$ & \citep{mmit, barnwal2022survival, rigaill:13} \\
\cmidrule(lr){1-6}
sum & 1 & 1 & 0 & $h_{t-1} + d_t$ & \citep{mmit, barnwal2022survival, rigaill:13} \\
\cmidrule(lr){1-6}
cumulative absolute & 1 & 1 & 0 & $h_{t-1} + |d_t|$ & \citep{mmit, barnwal2022survival, rigaill:13} \\
\cmidrule(lr){1-6}
cumulative squares & 1 & 1 & 0 & $h_{t-1} + d_t^2$ & \citep{mmit, barnwal2022survival, rigaill:13} \\
\cmidrule(lr){1-6}
difference sum & 2 & 1 & $0$ & $|d_t-d_{t-1}| + h_{t-1}$ & \citep{opart_pen_mlp, mmit, barnwal2022survival, rigaill:13} \\
\cmidrule(lr){1-6}
\multirow{2}{*}{value range} & \multirow{2}{*}{1} & \multirow{2}{*}{2} & $\infty$ & $\min(d_t, h_{t-1}^{(1)})$ & \multirow{2}{*}{\citep{opart_pen_mlp, mmit, barnwal2022survival, rigaill:13}} \\
                             &   &   & $-\infty$  & $\max(d_t, h_{t-1}^{(2)})$ & \\
\bottomrule
\end{tabular}
\caption{Formula of recurrent core for each feature approximation.}
\label{tab:feature_approximation}
\end{table}

\section{Novelty}
In previous work, the prediction of $\lambda$ involves two independent steps. 
In step 1, which manually extracts sequence features, redundancy may be introduced, and important hidden features may be missed. 
To address this, we propose a model that combines both steps into one, utilizing a recurrent network to automatically extract relevant features from the raw sequence to predict $\lambda$, see Figure \ref{fig:proposed}.

\paragraph{Feature Extraction via Recurrent Networks.} In a Recurrent Network with input size $k \geq 1$ and hidden size $m \geq 1$, processing a sequence $\mathbf{d} = [d_1, d_2, \dots, d_N] \in \mathbb{R}^N$, the hidden state at time $t$, called $h_t$, using a function $f$, is updated as:
\begin{equation} \label{eq:recurrnet}
    h_t = f(d_{[t-k+1:t]}; h_{t-1}) \in \mathbb{R}^m
\end{equation}
where $d_{[t-k+1:t]} = [d_{t-k+1}, d_{t-k+2}, ..., d_{t}] \in \mathbb{R}^k$.
Time stamp $t$ ranges from $k$ to $N$.
The initial hidden state $h_{k-1}$ can be assigned manually.
The final hidden state $h_N \in \mathbb{R}^m$ captures information from the entire sequence.
Thus, we use $h_N$ as the $m$ \textit{automatically extracted features} from the raw sequence.

\paragraph{$\lambda$ prediction.}
The prediction is the linear combination of $m$ extracted features:
\begin{equation*}
    \log \hat{\lambda} = h_N \cdot \beta + \beta_0 \text{\quad or \quad} \hat{\lambda} = e^{h_N \cdot \beta + \beta_0}
\end{equation*}
where $\beta \in \mathbb{R}^m$ and $\beta_0 \in \mathbb{R}$ are the parameters to be learned.

\paragraph{Feature Approximation Ability of Recurrent Networks.}
Recurrent networks have the ability to approximate sequence features.
The core component of these models is a neural network. 
For instance, the Vanilla Recurrent Network \citep{rnn} core is a fully connected neural network, which, according to the universal approximation theorem \citep{mlp}, can approximate any continuous function, regardless of its complexity, provided there are enough neurons in the hidden layer.
This is why, although these models may not directly produce exact statistical sequence features, their architecture enables them to approximate these features with a high degree of accuracy.
Refer to Table \ref{tab:feature_approximation} to see how recurrent networks are capable of approximating sequence features.

\begin{figure}[t]
\vskip 0.2in
\begin{center}
\centerline{\includegraphics[width=0.55\textwidth]{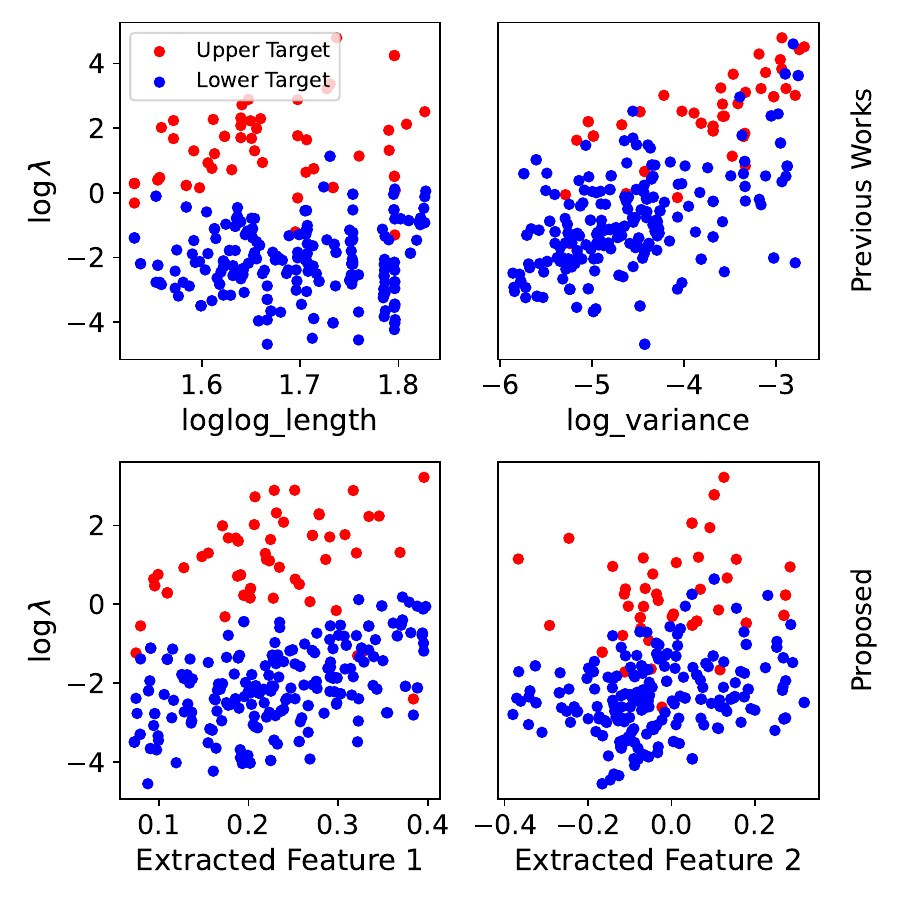}}
\caption{Example: features vs. targets from the dataset detailed.
The upper plots: the length or variance vs. targets, as these features are used on $\lambda$ prediction unsupervised methods.
The lower plots: automatically extracted features vs targets, using a GRU model with two extracted features.
Since we consider the predicted $\lambda$ has a linear relationship with the features, a good feature should exhibit a fairly linear pattern in the plot.}
\label{fig:extracted_features}
\end{center}
\vskip -0.2in
\end{figure}

\paragraph{Automatic vs. Manual features.} Figure \ref{fig:extracted_features} provides an example by comparing the ability to predict $\log\lambda$ using length (or variance) - as these features are considered key for $\log\lambda$ prediction in unsupervised methods - versus automatically extracted features from recurrent networks.
Since we assume that the predicted value of $\log\lambda$ has a linear relationship with the features, a good feature should display a linear pattern in the plot. 
It can be seen that the automatically extracted Feature 1 exhibits an excellent linear pattern with the all the target intervals since it's easy to visualize a line intersecting most of these intervals.

\paragraph{Recurrent network architectures.} This study explores 3 recurrent networks: Vanilla Recurrent Network (RNN), Long Short-Term Memory (LSTM) \citep{lstm}, Gated Recurrent Unit (GRU) \citep{gru}.
RNNs struggle with long-term dependencies due to the vanishing gradient problem. 
LSTMs address this by using memory cells and gating mechanisms that help retain information over longer sequences. 
GRUs, a simplified version of LSTMs, combine the forget and input gates into a single update gate, offering a more efficient alternative while still capturing long-term dependencies.

\begin{table}[t]
\caption{List of datasets with its number of sequences (Size)}
\label{tab:datasets}
\begin{small}
% \begin{sc}
\begin{tabular}{@{}lll|l@{}}
\toprule
\textbf{Dataset} & \textbf{Seq. length} & \textbf{Size} & \textbf{Source} \\
\midrule
systematic & \phantom{0000}66 - 5937 & 3418 & \\
detailed & \phantom{0000}25 - 5937 & 3730 & \href{https://github.com/tdhock/neuroblastoma-data/tree/master/data}{GitHub}\\ 
cancer & \phantom{0000}39 - 43628 & \phantom{0}826 & \\
\midrule
ATAC\_JV\_adipose & 105716 - 11.50M & \phantom{0}465 \\
CTCF\_TDH\_ENCODE & 166431 - 11.09M & \phantom{0}182 \\
H3K27ac-H3K4me3\_TDHAM\_BP & \phantom{000}275 -\phantom{0} 4.33M & 2008 \\
H3K27ac\_TDH\_some & \phantom{0}41053 -\phantom{0} 6.74M & \phantom{00}95 \\
H3K27me3\_RL\_cancer & \phantom{00}5730 -\phantom{0} 1.89M & \phantom{0}171 \\
H3K27me3\_TDH\_some & 113523 -\phantom{0} 3.00M & \phantom{00}43 \\
H3K36me3\_AM\_immune & 184092 -\phantom{0} 7.14M & \phantom{0}420 \\
H3K36me3\_TDH\_ENCODE & \phantom{0}68209 -\phantom{0} 2.74M & \phantom{00}78 \\
H3K36me3\_TDH\_immune & 434882 -\phantom{0} 4.35M & \phantom{00}37 & \href{https://archive.ics.uci.edu/ml/machine-learning-databases/00439/peak-detection-data.tar.xz}{UCI Repo}\\
H3K36me3\_TDH\_other & \phantom{0}49362 -\phantom{0} 7.15M & \phantom{00}40 \\
H3K4me1\_TDH\_BP & 148275 -\phantom{0} 9.73M & \phantom{0}144 \\
H3K4me3\_PGP\_immune & \phantom{0}24460 -\phantom{0} 6.62M & \phantom{0}297 \\
H3K4me3\_TDH\_ENCODE & \phantom{0}18198 -\phantom{0} 7.21M & \phantom{00}75 \\
H3K4me3\_TDH\_immune & \phantom{0}87557 -\phantom{0} 7.81M & \phantom{0}378 \\
H3K4me3\_TDH\_other & \phantom{0}80674 -\phantom{0} 3.72M & \phantom{00}90 \\
H3K4me3\_XJ\_immune & \phantom{0}52498 -\phantom{0} 5.50M & \phantom{0}270 \\
H3K9me3\_TDH\_BP & 369654 - 10.29M & \phantom{0}120 \\
\bottomrule
\end{tabular}
% \end{sc}
\end{small}
\end{table}

\section{Experiments} \label{experiment}
In this section, the implementation of models is detailed. 
Labeled sequences are first processed, with features manually extracted to replicate previous work, along with the corresponding target intervals. 
Baseline models and the proposed models are then implemented. 
For each test sequence, OPART is applied with the predicted $\lambda$ to detect changepoints. 
Finally, accuracy rates are calculated based on the predicted changepoints and the test set labels.

\paragraph{Raw Sequence Datasets.}  
This study uses 20 datasets: two DNA copy number profiles from neuroblastoma tumors \citep{rigaill:13}, known for detailed and systematic dataset; one dataset cancer from \citep{review:lopart}; and 17 ChIP-seq datasets from \citep{hocking_peak16}, see Table \ref{tab:datasets} for all benchmark datasets, each sequence has a different length (number of datapoints), so the range of sequence lengths within each dataset is reported in the Table \ref{tab:datasets}.
For each sequence in each dataset, biologist experts use their visual interpretation of important signal and noise patterns to label regions as having changes in mean value or not.

\paragraph{Train/Test Setup.}  
Each sequence is assigned a unique identifier (sequenceID), and datasets are split into folds based on these IDs. Fewer folds are used for smaller datasets to balance train and test data sizes, ensuring a sufficiently large test set for reliable evaluation. In each iteration, one fold serves as the test set, and the others form the train set.

\paragraph{Evaluation Metrics.}
The metric used is the accuracy rate on the test set:
\[
\text{Accuracy} = \frac{TP + TN}{TP + TN + FP + FN}
\]
where
\begin{itemize}
    \item TP = \text{True Positives (The number of detected changepoints matches to the positive labels)}
    \item TN = \text{True Negatives (No detected changepoints in the negative labels)}
    \item FP = \text{False Positives (The number of detected changepoints exceeds that of the labels.)}
    \item FN = \text{False Negatives (The number of detected changepoints is less than that of the labels.)}
\end{itemize}

\subsection{Model Implementation}
Previous models were implemented as baselines, along with additional models and the proposed models for comparison. 
Details of all models are provided in Table \ref{tab:models}.

\begin{table}[t]
\caption{List of employed models}
\label{tab:models}
\begin{small}
% \begin{sc}
\begin{tabular}{@{}lllll@{}}
\toprule
\textbf{Model} & \textbf{Regularization} & \textbf{Language} & \textbf{Package} & \textbf{Paper} \\
\midrule
linear & L1 & R & \citep{penaltyLearning} & \cite{rigaill:13} \\
\midrule
MMIT & tree architecture & Python/Cpp binding &  \citep{mmit_R} & \cite{mmit} \\
\midrule
AFT\_XGboost & tree architecture & Python & \citep{xgboost_python} & \cite{barnwal2022survival} \\
\midrule
constant & none & Python/Cpp binding &  \citep{mmit_R} & Baseline \\
\midrule

\begin{tabular}[c]{@{}l@{}}MLP\end{tabular} & \begin{tabular}[c]{@{}l@{}}hidden layer number\\ hidden layer sizes\end{tabular} & Python &  & Additional \\
\midrule
\begin{tabular}[c]{@{}l@{}}RNN\\LSTM\\GRU\end{tabular} & \begin{tabular}[c]{@{}l@{}}hidden layer number\\ hidden state sizes\end{tabular} & Python & & Proposed \\
\bottomrule
\end{tabular}
% \end{sc}
\end{small}
\end{table}

\subsubsection{Baseline Models}
\paragraph{Linear.} This model uses L1 regularization. 
The regularization parameter L1 starts at 0.001, geometrically increasing by a factor of 1.2 until no features remain, with cross-validation determining the optimal L1 regularization value.

\paragraph{MMIT.}  The optimal hyperparameters are chosen through cross-validation on the training set, including:
\begin{itemize}
    \item \texttt{max\_depth}:  0, 1, 5, 10, 20, and $\infty$
    \item \texttt{min\_sample}:  0, 1, 2, 4, 8, 16, and 20. 
    \item \texttt{margin} ($\epsilon$ in Hinge Error \ref{eq:loss_function}):  0, 1, and 2
    \item \texttt{loss\_type}:  \texttt{hinge} and \texttt{square} (equivalent to $p=1$ and $p=2$ in Hinge Error \ref{eq:loss_function});
\end{itemize}

\begin{figure}[t]
\vskip 0.2in
\begin{center}
\centerline{\includegraphics[width=0.8\textwidth]{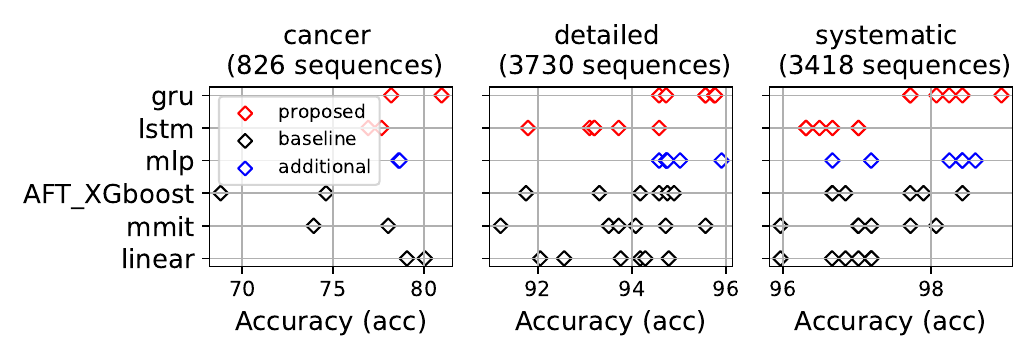}}
\caption{The accuracies for each fold are presented for datasets without any pooling preprocessing. The cancer dataset has 2 folds, while the other two datasets each have 6 folds. The Constant and RNN models have been excluded due to their small accuracies to better highlight the differences and improve visualization. Overall, the GRU model achieves the highest accuracy.}
\label{fig:acc_no_pooling}
\end{center}
\vskip -0.2in
\end{figure}

\paragraph{AFT in XGBoost.} Following the same setup described in \citet{barnwal2022survival}, the optimal hyperparameters are chosen via cross-validation on the training set, including:
\begin{itemize}
    \item \texttt{learning\_rate}: 0.001, 0.01, 0.1, and 1.0
    \item \texttt{max\_depth}: 2, 3, 4, 5, 6, 7, 8, 9, and 10
    \item \texttt{min\_child\_weight}: 0.001, 0.1, 1.0, 10.0, and 100.0
    \item \texttt{reg\_alpha}: 0.001, 0.01, 0.1, 1.0, 10.0, and 100.0
    \item \texttt{reg\_lambda}: 0.001, 0.01, 0.1, 1.0, 10.0, and 100.0
    \item \texttt{aft\_loss\_distribution\_scale}: 0.5, 0.8, 1.1, 1.4, 1.7, and 2.0
\end{itemize}

\paragraph{Constant (Featureless).} 
This model predicts a single value $\hat{\lambda} = \exp{\log\hat{\lambda}}$ from the train set using only target intervals.
The objective is to minimize the total Hinge Error \eqref{eq:loss_function} over $ S $ instances with target intervals $[\log\lambda_l^i, \log\lambda_u^i]_{i=1}^S$:  
\begin{equation*}
\log\hat{\lambda} = \argmin_{c} \sum_{i=1}^{S} l(c, [\log\lambda_l^i, \log\lambda_u^i]).
\end{equation*}

\subsubsection{Additional Model}
\paragraph{MLP.}
Using the same feature vector and loss function as the linear model, the MLP generalizes the linear model by incorporating ReLU activation in the hidden layers.
Hyperparameters include the number of hidden layers (1, 2, or 3) and the size of each hidden layer (1, 2, 4, 8, 16, 32, 64, 128, 256, or 512).
The best hyperparameters are chosen via cross-validation in the train set.
Since the linear model uses L1 regularization and MMIT (or AFT in XGBoost) has a tree-based architecture, they can automatically select features. 
However, MLP does not fully mitigate irrelevant features on its own (parameters associated with each feature are likely non-zero). 
Feature selection for MLP in this interval regression setting could be explored in future work.

\begin{figure}[t]
\vskip 0.2in
\begin{center}
\centerline{\includegraphics[width=0.8\textwidth]{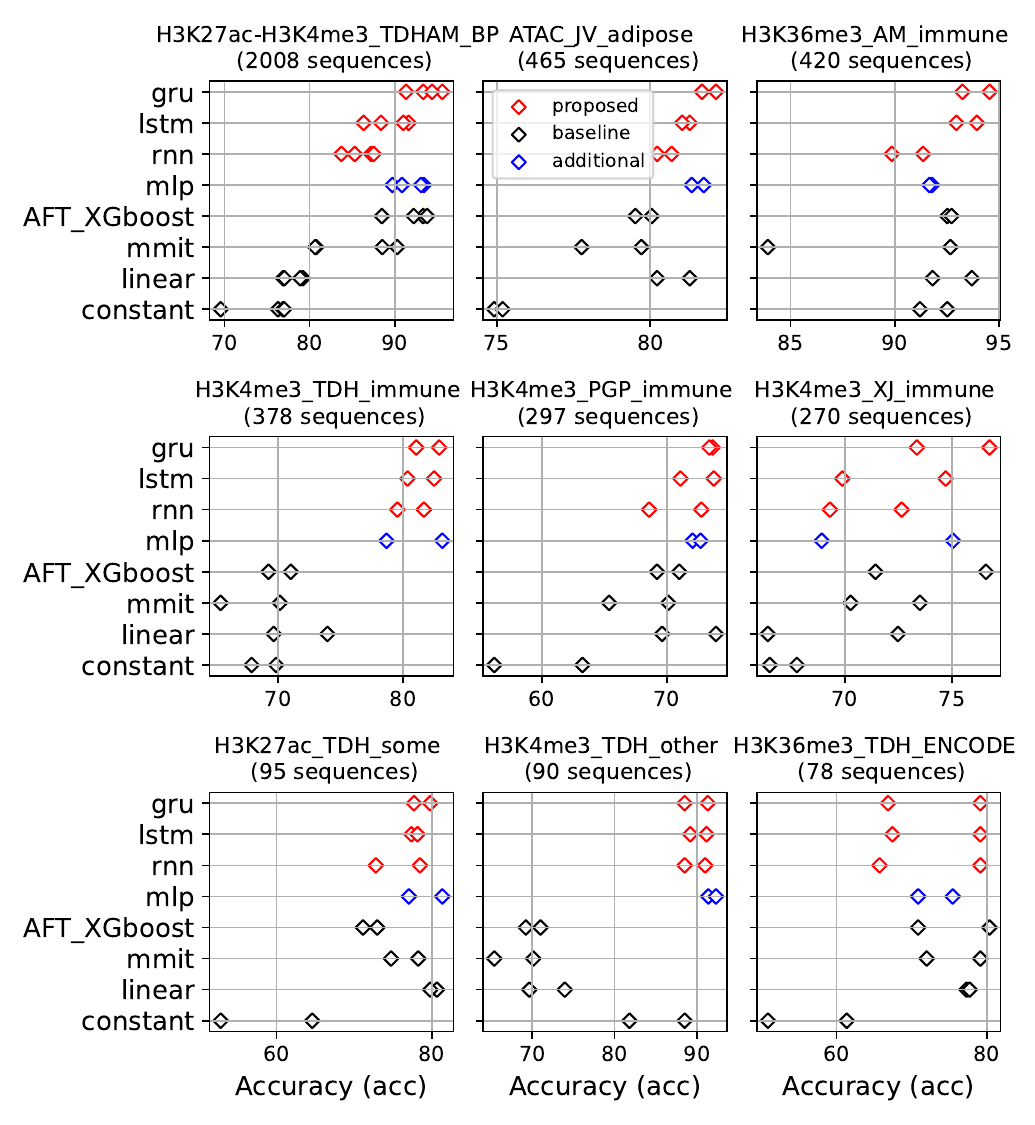}}
\caption{The fold accuracies on some ChIP-seq datasets are presented, with the datasets ordered from largest to smallest. 
The number of colored diamonds in any row represents the number of folds for each dataset. As observed, for larger datasets, more complex models tend to show higher accuracy. For the last three smaller datasets, recurrent networks do not always outperform the previous methods.}
\label{fig:acc_chipseq}
\end{center}
\vskip -0.2in
\end{figure}

\subsubsection{Proposed Models}
\paragraph{Architecture.} There are three considered models: RNN, LSTM, and GRU. The hyperparameters include: the number of hidden layers (1 or 2) and the hidden state size (number of extracted features) (2, 4, 8 or 16).

\paragraph{Preprocessing.} Based on the datasets sequence lengths, we handle them as follows:
\begin{itemize}
    \item For the 3 datasets—detailed, systematic, cancer—the sequence lengths range from 39 to 43628. 
    These sequences are directly input into the $\lambda$ prediction models without further modification.
    \item For the 17 ChIP-seq datasets, the sequence lengths can be over 11 million.
    Training on these raw sequences can be computationally expensive and time-consuming.
    So pooling operations are performed, where a single representative value is chosen for each non-overlapping sliding window. The window size options are 100, 1000, or 2000, and the representative value is either the window mean or the median. After shortening the sequences, a $\log(z+1)$ transformation is applied to all value $z$ in all shortened sequences to mitigate skewed distributions. Following this, all sequence values are normalized (mean 0, variance 1) for each dataset.
\end{itemize}

\subsection{Results}
The comprehensive accuracy results of all 8 methods for each of 20 datasets are provided in the Appendix, as the table is too large to be included in the main paper.
Figure \ref{fig:acc_no_pooling} presents the accuracy rates for three datasets using the original, unpooled sequences, while Figure \ref{fig:acc_chipseq} shows the results for various ChIP-seq datasets (pooled sequences).
Each dataset is evaluated using seven different methods, producing multiple accuracy values depending on the number of folds used for that dataset.

\section{Discussion and Conclusion}
\paragraph{Summary of Contributions.}
Technically, these proposed models still perform two steps: (1) Feature extraction and (2) Predicting $\lambda$ based on the extracted features.
However, instead of performing these steps separately, it combines them into a single operation, enhancing its robustness compared to previous models.
Automatic extracting features helps generate highly useful ones without requiring many features, as shown in Figure \ref{fig:extracted_features}.
This study focuses exclusively on genomic datasets, where prior knowledge of relevant features enables effective manual feature extraction. 
However, the proposed models stand out because it does not require expert knowledge to perform. Its automatic feature extraction mechanism makes it versatile and applicable to any type of sequence dataset.

\paragraph{Comparison to Previous Models.}
Recurrent networks, in general, outperform previous models across datasets, see Figure \ref{fig:acc_no_pooling} and Figure \ref{fig:acc_chipseq}.
For small datasets (with fewer than 100 instances), the accuracy of the proposed models is comparable (neither better nor worse) to that of existing models. 
On the other hand, when dealing with larger datasets, recurrent networks noticeably outperform previous models and show a slight edge over MLPs.

\paragraph{Comparison between Recurrent Networks.}
When comparing RNN, LSTM, and GRU, as expected, RNN performs the worst due to its difficulty with long-term memory. 
In most cases, GRU outperforms the others.

\begin{figure}[t]
\vskip 0.2in
\begin{center}
\centerline{\includegraphics[width=0.7\textwidth]{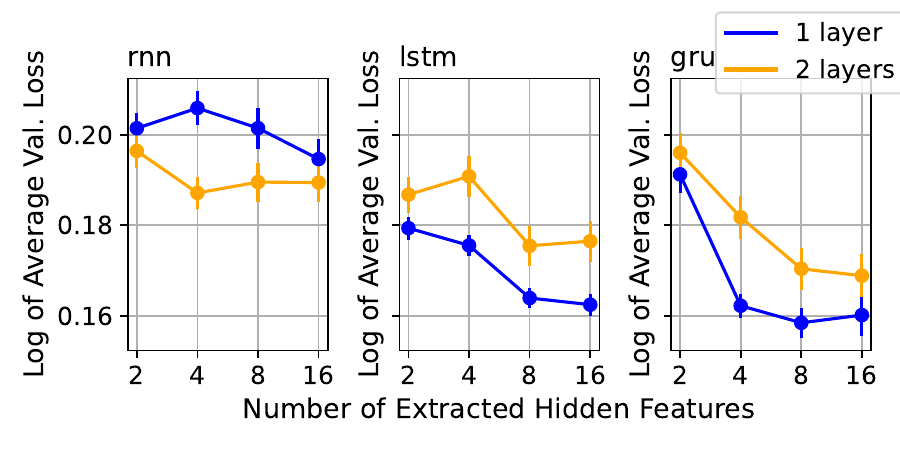}}
\caption{Log of the loss validation with 68.27\% confidence interval across different numbers of extracted hidden layers in the three proposed models shows that the number of extracted hidden features needs to be sufficient. As seen in the figure, when the number of features is 8 or 16, the loss value is lower compared to when the number of features is 2 or 4.}
\label{fig:n_feature}
\end{center}
\vskip -0.2in
\end{figure}

\paragraph{Discussion on the Number of Automatically Extracted Features.}  
As shown in the Figure \ref{fig:n_feature}, the number of automatic extracted features needs to be sufficiently large. 
It is observed that when 8 or 16 features are extracted, the validation loss value is slightly smaller compared to when only 2 or 4 features are extracted.

\begin{figure}[t]
\vskip 0.2in
\begin{center}
\centerline{\includegraphics[width=0.7\textwidth]{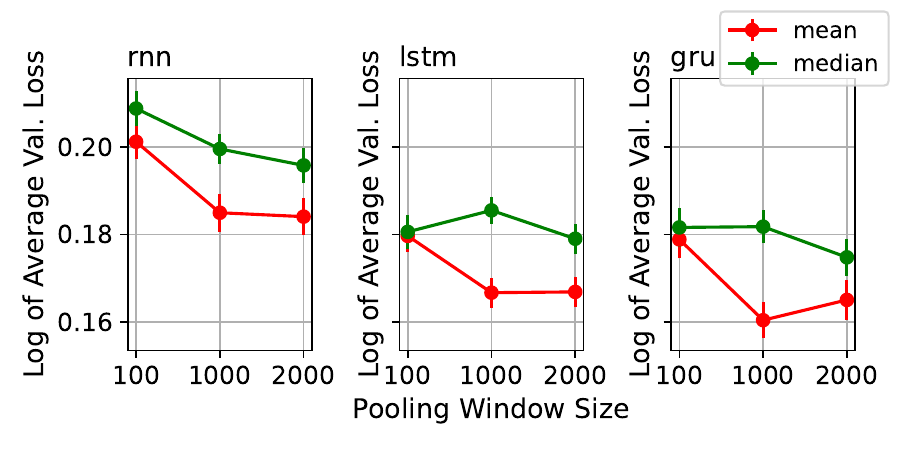}}
\caption{Log of the loss validation with 68.27\% confidence interval across different pooling methods in the three proposed models reveals that the window size must be large enough to reduce training time and difficulty, yet small enough to retain crucial information. The figure shows that mean pooling outperforms median pooling. These results are based on 17 ChIP-seq datasets.}
\label{fig:window_size}
\end{center}
\vskip -0.2in
\end{figure}

\paragraph{Impact of pooling preprocessing.}
Figure \ref{fig:window_size} shows that shortening the sequence for training is effective. As the window size increases, more information is lost. 
However, when the window size is small, such as 100, the performance of the recurrent models is worse compared to larger window sizes like 1000 or 2000. This could be because the pooled sequence with a window size of 100 is still too long for the recurrent models to handle effectively. 
The pooling process is beneficial in two ways: it can improve the accuracy of recurrent networks and also reduce training time.
We employed two pooling functions, mean and median, to represent a single value for each non-overlapping window. 
As shown in Figure \ref{fig:window_size}, it is evident that using the mean value as the representation consistently yields smaller loss.

\paragraph{Study Limitations.}
Although the Recurrent Network architecture is straightforward, as its core essentially a deep neural network, training these models can be time-consuming due to the inherently sequential nature of the process.
For some train/test pairs, even after applying pooling to reduce the sequence length, training a model with a maximum of 2 layers, each containing no more than 16 neurons, and a maximum of 1000 iterations, took over approximately 10 days, despite utilizing the powerful NVIDIA A100 GPU.
These models also require significant memory to store the hidden states at each time step. 
To avoid out-of-memory (OOM) errors, it is crucial to allocate sufficient memory, especially for long sequences.

\paragraph{Practical Considerations in Real-World Applications.}
There is a trade-off between training time and accuracy. 
The proposed models have higher accuracy in most of the 20 datasets, but they are significantly more time-consuming to train, especially for longer sequences. 
In contrast, manually extracted features models do not face this issue.
Therefore, in real-world scenarios where datasets contain only a few sequences but each sequence is long, and training time is a priority, manually extracted features models may be the better choice for two reasons: first, their training time is significantly faster; and second, these models' accuracy is comparable to the proposed models, which can be seen in the bottom three panels of Figure~\ref{fig:acc_chipseq}.
However, if the user’s priority is accuracy and the hardware requirements are met, the proposed models are a better choice.
Although the training process is time-consuming, it only needs to be done once; afterwards, the trained models can be applied to unseen sequences, as their feed-forward process is sufficiently fast.

\newpage
\paragraph{Future Directions.}
For these particular datasets, it is assumed that $\lambda$ exhibits an increasingly monotonic behavior with respect to both sequence length and variance, which are employed in unsupervised methods. 
One approach is to leverage monotonic network architectures, such as min-max networks \citep{monotonic}, or more advanced architectures like Lattice Networks \citep{lattice}, with modifications to the loss function for each linear region.

We can consider CNNs \citep{cnn} because they are capable of handling sequences of varying lengths and generating a single output by utilizing convolution layers to extract features across the sequence. 
Global pooling layers (e.g., Global Max or Average Pooling) aggregate these features into a fixed-size representation. This vector is then passed through a MLP to produce the final output.

To build upon the initial concept of utilizing recurrent networks in this study, various types of models could be explored in the future:
\begin{itemize}
    \item Using a more complex model: This study predicts $\lambda$ by using a linear combination of the extracted features. Alternatively, we could consider using a more complex model such as MLP that processes the extracted features instead of relying on a linear combination.
    \item Attention mechanisms \citep{attention} dynamically focus on relevant input parts, assigning importance weights to subsequences. Unlike sequential RNNs, they effectively capture long-range dependencies and handle varying input lengths, making them ideal for this task.
    \item Bidirectional Recurrent Networks \citep{bidirectional}: The models implemented in this study are unidirectional, leading to challenges with long-term memory, since the later parts of the sequence tend to be more important than the earlier ones. Using a bidirectional recurrent network architecture can aim to address this limitation, enabling the model to better capture information from the sequence.
\end{itemize}
Conducting even a preliminary experiment would require substantial computing resources, since bidirectional networks and attention mechanisms are more computationally expensive than the proposed one. For this reason, we chose not to include such experiments as part of this work.

\section*{Funding sources}
This research did not receive any specific grant from funding agencies in the public, commercial, or not-for-profit sectors.

\newpage
\bibliography{ref}

\newpage
\section*{Appendix}
% \subsection*{Accuracy statistics of each dataset with respect to each method}
\begin{sidewaystable}
\footnotesize
\caption{Accuracy means and standard deviations (by method)} 
\begin{tabular}{|lllllllll|}
\hline
\multicolumn{1}{|c}{\textbf{Dataset}} & \multicolumn{1}{c}{\textbf{constant}} & \multicolumn{1}{c}{\textbf{linear}} & \multicolumn{1}{c}{\textbf{mmit}} & \multicolumn{1}{c}{\textbf{AFT\_XGB}} & \multicolumn{1}{c}{\textbf{mlp}}   & \multicolumn{1}{c}{\textbf{rnn}}   & \multicolumn{1}{c}{\textbf{lstm}}  & \multicolumn{1}{c|}{\textbf{gru}}   \\ \hline
\textbf{cancer}                        & 63.37 ± 14.93                          & 79.55 ± 0.70                         & 75.97 ± 2.90                       & 71.70 ± 4.10                               & 78.62 ± 0.04                        & 75.42 ± 2.12                        & 77.29 ± 0.52                        & {\color[HTML]{FE0000} 79.58 ± 1.96} \\ \hline
\textbf{detailed}                      & 67.45 ± 6.42                           & 93.60 ± 1.07                         & 93.80 ± 1.47                       & 93.91 ± 1.20                               & 94.95 ± 0.49                        & 86.26 ± 4.64                        & 93.26 ± 0.91                        & {\color[HTML]{CB0000} 95.32 ± 0.53} \\ \hline
\textbf{systematic}                    & 56.87 ± 9.87                           & 96.81 ± 0.46                         & 97.16 ± 0.72                       & 97.37 ± 0.75                               & 97.92 ± 0.80                        & 72.87 ± 29.01                       & 96.64 ± 0.32                        & {\color[HTML]{CB0000} 98.19 ± 0.47} \\ \hline
\textbf{ATAC\_JV\_adipose}             & 75.05 ± 0.19                           & 80.75 ± 0.75                         & 78.74 ± 1.38                       & 79.78 ± 0.38                               & 81.55 ± 0.28                        & 80.46 ± 0.33                        & 81.16 ± 0.17                        & {\color[HTML]{CB0000} 81.91 ± 0.32} \\ \hline
\textbf{CTCF\_TDH\_ENCODE}             & 75.47 ± 1.71                           & 84.15 ± 0.40                         & 87.65 ± 0.54                       & 93.32 ± 0.84                               & {\color[HTML]{CB0000} 93.96 ± 0.38} & 86.58 ± 0.23                        & 89.97 ± 2.50                        & 91.52 ± 7.80                        \\ \hline
\textbf{H3K27ac-H3K4me3\_TDHAM\_BP}    & 74.93 ± 3.60                           & 78.00 ± 1.22                         & 85.06 ± 5.06                       & 91.95 ± 2.41                               & 91.76 ± 1.77                        & 85.94 ± 1.78                        & 89.33 ± 2.44                        & {\color[HTML]{CB0000} 93.65 ± 1.80} \\ \hline
\textbf{H3K27ac\_TDH\_some}            & 58.69 ± 8.35                           & {\color[HTML]{CB0000} 80.23 ± 0.60}  & 76.51 ± 2.48                       & 72.06 ± 1.30                               & 79.21 ± 3.05                        & 75.63 ± 4.03                        & 77.77 ± 0.56                        & 78.76 ± 1.49                        \\ \hline
\textbf{H3K27me3\_RL\_cancer}          & 54.86 ± 1.35                           & 64.52 ± 5.54                         & 65.90 ± 2.32                       & 65.23 ± 0.78                               & {\color[HTML]{CB0000} 67.58 ± 3.57} & 64.43 ± 3.32                        & 65.49 ± 3.24                        & 64.22 ± 4.08                        \\ \hline
\textbf{H3K27me3\_TDH\_some}           & 62.19 ± 2.89                           & 85.48 ± 1.90                         & 77.19 ± 6.68                       & 85.40 ± 3.35                               & 77.83 ± 5.26                        & 85.43 ± 2.36                        & {\color[HTML]{CB0000} 87.21 ± 2.26} & 86.66 ± 2.01                        \\ \hline
\textbf{H3K36me3\_AM\_immune}          & 91.85 ± 0.93                           & 92.75 ± 1.33                         & 88.29 ± 6.18                       & 92.61 ± 0.15                               & 91.72 ± 0.09                        & 90.60 ± 1.06                        & 93.44 ± 0.70                        & {\color[HTML]{CB0000} 93.89 ± 0.92} \\ \hline
\textbf{H3K36me3\_TDH\_ENCODE}         & 56.10 ± 7.42                           & {\color[HTML]{CB0000} 77.51 ± 0.29}  & 75.57 ± 5.05                       & 75.61 ± 6.73                               & 73.16 ± 3.25                        & 72.43 ± 9.49                        & 73.28 ± 8.28                        & 73.00 ± 8.69                        \\ \hline
\textbf{H3K36me3\_TDH\_immune}         & {\color[HTML]{CB0000} 96.39 ± 1.96}    & 87.78 ± 11.00                        & 87.63 ± 0.32                       & 90.13 ± 3.85                               & 88.62 ± 8.01                        & 95.67 ± 2.97                        & 95.67 ± 2.97                        & 95.67 ± 2.97                        \\ \hline
\textbf{H3K36me3\_TDH\_other}          & 77.00 ± 2.83                           & 93.50 ± 2.12                         & 79.00 ± 15.56                      & 91.50 ± 0.71                               & 70.50 ± 17.68                       & 94.09 ± 2.28                        & {\color[HTML]{CB0000} 94.09 ± 2.28} & 88.74 ± 5.28                        \\ \hline
\textbf{H3K4me1\_TDH\_BP}              & 78.63 ± 1.69                           & 89.10 ± 2.37                         & 87.02 ± 2.18                       & 85.20 ± 1.68                               & {\color[HTML]{CB0000} 89.43 ± 1.56} & 88.39 ± 1.63                        & 88.79 ± 1.07                        & 88.39 ± 1.63                        \\ \hline
\textbf{H3K4me3\_PGP\_immune}          & 59.73 ± 5.00                           & 71.79 ± 3.04                         & 67.76 ± 3.35                       & 70.11 ± 1.27                               & 72.39 ± 0.45                        & 70.68 ± 2.95                        & 72.43 ± 1.89                        & {\color[HTML]{CB0000} 73.55 ± 0.18} \\ \hline
\textbf{H3K4me3\_TDH\_ENCODE}          & 92.33 ± 1.83                           & 94.78 ± 1.04                         & 95.37 ± 1.74                       & {\color[HTML]{CB0000} 97.05 ± 1.25}        & 93.89 ± 2.79                        & 94.60 ± 1.62                        & 94.94 ± 1.14                        & 94.94 ± 1.14                        \\ \hline
\textbf{H3K4me3\_TDH\_immune}          & 68.85 ± 1.39                           & 71.79 ± 3.04                         & 67.76 ± 3.35                       & 70.11 ± 1.27                               & 80.91 ± 3.16                        & 80.61 ± 1.50                        & 81.42 ± 1.50                        & {\color[HTML]{CB0000} 81.98 ± 1.29} \\ \hline
\textbf{H3K4me3\_TDH\_other}           & 85.15 ± 4.73                           & 71.79 ± 3.04                         & 67.76 ± 3.35                       & 70.11 ± 1.27                               & {\color[HTML]{CB0000} 91.83 ± 0.65} & 89.74 ± 1.76                        & 90.16 ± 1.39                        & 89.90 ± 1.99                        \\ \hline
\textbf{H3K4me3\_XJ\_immune}           & 67.13 ± 0.89                           & 69.44 ± 4.29                         & 71.89 ± 2.29                       & 74.00 ± 3.64                               & 71.98 ± 4.33                        & 70.98 ± 2.36                        & 72.29 ± 3.41                        & {\color[HTML]{CB0000} 75.06 ± 2.40} \\ \hline
\textbf{H3K9me3\_TDH\_BP}              & 48.94 ± 1.50                           & 84.29 ± 1.75                         & 85.94 ± 0.11                       & 79.39 ± 6.81                               & 84.19 ± 2.81                        & {\color[HTML]{CB0000} 86.16 ± 1.43} & 85.04 ± 1.62                        & 86.08 ± 2.48                        \\ \hline

\end{tabular}
\label{tab:yourlabel}
\end{sidewaystable}

\end{document}